\documentclass[runningheads]{llncs}
\usepackage[T1]{fontenc}
\usepackage{graphicx,verbatim}
\usepackage{multirow}
\usepackage{booktabs}
\usepackage{threeparttable}
\usepackage{pifont}
\usepackage{amsmath}
\usepackage{bm}
\usepackage{enumitem,amssymb}
\usepackage[misc]{ifsym}

\newcommand{\ifmri}{\mathbf{X}_{raw}}
\newcommand{\ifc}{\mathbf{X}_{C}}
\newcommand{\its}{\mathbf{X}_{T}}
\newcommand{\rfc}{\mathbf{R}_{C}}
\newcommand{\rts}{\mathbf{R}_{T}}
\newcommand{\rfinal}{\mathbf{R}_{TC}}
\newcommand{\efc}{E_{C}}
\newcommand{\ets}{E_{T}}
\newcommand{\ourmodel}{FMM$_{TC}$}

\newcommand{\cmark}{\ding{51}}%
\newcommand{\xmark}{\ding{55}}%

\newcommand{\corrauth}{\textsuperscript{(\Letter)}}
\usepackage{hyperref}
\usepackage{color}

\begin{document}
\title{Foundation-Model-Boosted Multimodal Learning for fMRI-based Neuropathic Pain Drug Response Prediction}
\titlerunning{Foundation-Model-Boosted Multimodal Learning}
\author{Wenrui Fan\inst{1,2}\orcidID{0009-0007-1394-0092} \and
L. M. Riza Rizky\inst{1,2}\orcidID{0009-0003-3073-9789} \and
Jiayang Zhang\inst{1,2}\orcidID{0009-0007-8446-2462} \and
Chen Chen\inst{1,2,6}\orcidID{0000-0002-3525-9755} \and
Haiping Lu\inst{1,2}\orcidID{0000-0002-0349-2181} \and
Kevin Teh\inst{3}\orcidID{0000-0003-2538-5157} \and
Dinesh Selvarajah\inst{4,5}\orcidID{0000-0001-7426-1105} \and
Shuo Zhou\inst{1,2}\corrauth\orcidID{0000-0002-8069-2814}
}
\authorrunning{W. Fan et al.}
%
\institute{
School of Computer Science, University of Sheffield, Sheffield, UK \and
Centre for Machine Intelligence, University of Sheffield, Sheffield, UK \and
Academic Unit of Radiology, University of Sheffield, Sheffield, UK \and
Diabetes Research Unit, Royal Hallamshire Hospital, Sheffield, UK \and
Division of Clinical Medicine, University of Sheffield, Sheffield, UK \and
Department of Computing, Imperial College London, London, UK\\
\email{\{wenrui.fan, l.m.rizky, jiayang.zhang, chen.chen2, h.lu, k.teh, d.selvarajah, shuo.zhou\corrauth\}@sheffield.ac.uk}}


\maketitle              
\begin{abstract}
Neuropathic pain, affecting up to 10\% of adults, remains difficult to treat due to limited therapeutic efficacy and tolerability. 
Although resting-state functional MRI (rs-fMRI) is a promising non-invasive measurement of brain biomarkers to predict drug response in therapeutic development, the complexity of fMRI demands machine learning models with substantial capacity.
However, \emph{extreme data scarcity} in neuropathic pain research limits the application of high-capacity models.
To address the challenge of data scarcity, we propose \textbf{\ourmodel}, a \textbf{F}oundation-\textbf{M}odel-boosted \textbf{M}ultimodal learning framework for fMRI-based neuropathic pain drug response prediction, which leverages both internal multimodal information in \emph{pain-specific} data and external knowledge from large \emph{pain-agnostic} data.
Specifically, to maximize the value of limited pain-specific data, \ourmodel\ integrates complementary information from two rs-fMRI modalities: \textbf{T}ime series and functional \textbf{C}onnectivity.
\ourmodel\ is further boosted by an fMRI foundation model with its external knowledge from extensive pain-agnostic fMRI datasets enriching limited pain-specific information.
Evaluations with an in-house dataset and a public dataset from OpenNeuro demonstrate \ourmodel's superior representation ability, generalizability, and cross-dataset adaptability over existing unimodal fMRI models that only consider one of the rs-fMRI modalities.
The ablation study validates the effectiveness of multimodal learning and foundation-model-powered external knowledge transfer in \ourmodel.
An integrated gradient-based interpretation study explains how \ourmodel's cross-dataset dynamic behaviors enhance its adaptability.
In conclusion, \ourmodel\ boosts clinical trials in neuropathic pain therapeutic development by accurately predicting drug responses to improve the participant stratification efficiency.
The code is released on \url{https://github.com/Shef-AIRE/FMM_TC}.


\end{abstract}

\section{Introduction}

Neuropathic pain, caused by damage or disease in the somatosensory nervous system~\cite{def1}, affects up to 10\% of people over 30 years old~\cite{ten-percent2}. 
Despite advances in pharmacological treatments, nearly half of patients fail to achieve adequate pain relief~\cite{bad-treat2} due to low drug efficacy~\cite{bad-treat1} and high rates of side effects~\cite{side-effect1}. 
This underscores the urgent need for novel therapeutic approaches~\cite{ai4pain-review2}.

Although resting-state functional magnetic resonance imaging (rs-fMRI) is a promising non-invasive measurement of brain biomarkers~\cite{ai4pain-review2,fmri-drug}, its high dimensionality~\cite{complexity}, noisiness~\cite{noisy}, and cross-individual variability~\cite{variability} demand machine learning (ML) models with substantial capacity. 
To capture complex patterns in fMRI data, previous works have proposed advanced deep neural networks with strong representation ability (e.g. transformers~\cite{bnt} and Mamba-based models~\cite{mamba}) that perform well with large-scale training data such as UK Biobank~\cite{ukb2}.

However, directly applying such cutting-edge machine learning models for neuropathic pain fMRI analysis faces a significant challenge of \emph{extreme data scarcity}, leading to severe overfitting or training collapse problems.
In fact, neuropathic pain datasets usually have less than 100 participants~\cite{openneuro} due to the challenges in the data acquisition process (e.g. high costs, ethics).
Therefore, practical applications of advanced machine learning models with sophisticated network architectures for neuropathic pain fMRI analysis are hindered. 

\begin{figure}[t]
    \centering
    \includegraphics[width=\textwidth]{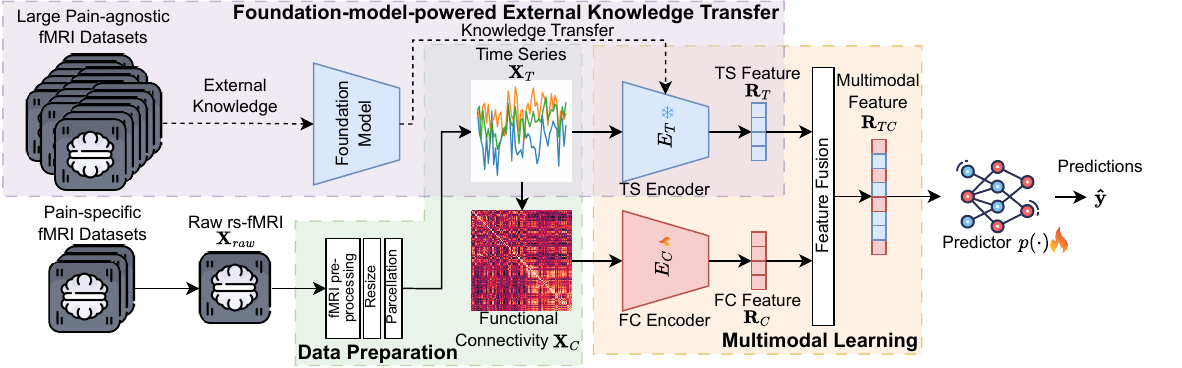}
    \caption{
    \ourmodel\ pipeline. The raw rs-fMRI $\ifmri$ undergoes three processing steps to obtain time-series (TS) $\its$.
    Then, functional connectivity (FC) $\ifc$ is computed from $\its$.
    Next, $\its$ and $\ifc$ are processed by a time-series (TS) encoder $\ets$ (frozen fMRI foundation model) and a functional connectivity (FC) encoder $\efc$ (learnable CNN), respectively. 
    Finally, features of two modalities $\rts$ and $\rfc$ are fused to get the multimodal feature $\rfinal$ for the following prediction.
    }
    \label{fig:main}
\end{figure}

To tackle this challenge, we propose \textbf{\ourmodel}, a \textbf{F}oundation-\textbf{M}odel-boosted \textbf{M}ultimodal learning framework for fMRI-based drug response prediction in neuropathic pain therapeutic development. 
As shown in Fig.~\ref{fig:main}, \ourmodel\ proposes two novel modules to address data scarcity: a) multimodal learning across two modalities of rs-fMRI: \textbf{T}ime series (TS) and functional \textbf{C}onnectivity (FC), and b) transferring external knowledge from large-scale pain-agnostic fMRI datasets via foundation models (FMs).
Our contributions are three-fold:

\emph{Firstly}, we introduce multimodal learning in \ourmodel\ to exploit internal information in the limited pain-specific data by integrating complementary information from TS and FC modalities.
Although previous works~\cite{bnt,brainnetcnn,brainlm} regard FC and TS as one modality due to coming from the same source, \ourmodel\ rethinks them as two distinct modalities, encodes them in two separate streams, and fuses their latent features by a simple yet effective design to enhance predictions.
By multimodal learning, \ourmodel\ improves Matthews correlation coefficient (MCC) by at least 2.80\% compared to unimodal baselines across all experiments.

\emph{Secondly}, \ourmodel\ exploit the benefits of external knowledge that FM~\cite{brainlm} learns from large-scale pain-agnostic fMRI datasets (e.g. UK Biobank~\cite{ukb2} with 50,000+ scans).
Enriching limited internal information with rich knowledge from the general population~\cite{ukb2} beyond pain patients, this FM-powered external knowledge transfer improves MCC by at least 14.71\% compared to the model with the same network architecture but no pre-training or external knowledge.

\emph{Thirdly}, we comprehensively evaluate \ourmodel\ with one public dataset from OpenNeuro~\cite{openneuro} and an in-house dataset.
Drug-agnostic and within-domain drug-specific response prediction demonstrate \ourmodel's promising representation ability and cross-dataset adaptability (up to MCC=69.64\% and AUROC=77.56\%).
The out-of-domain drug-specific response prediction results show \ourmodel's strong generalizability (up to MCC=86.02\%).
We conduct an ablation study to demonstrate the effectiveness of multimodal learning and external knowledge transfer, and an interpretation study to understand the reason for its strong adaptability.

\section{Methodology}

\noindent\textbf{Problem formulation.}
The drug response prediction task is formulated as a binary classification problem. 
With a dataset of rs-fMRI images $\{\ifmri\}$ from a cohort of participants, the goal is to learn a function $f_{\bm{\theta}}$ parameterized by $\bm{\theta}$ to predict participant's drug response $\hat{y}=f_{\bm{\theta}}(\ifmri)$ with label $y$=1 indicating effective (responder) and 0 indicating no response (non-responder).

\noindent\textbf{Overview of \ourmodel.} \ourmodel\ has three major modules: a) data preparation, b) multimodal learning, and c) FM-powered external knowledge transfer.

\noindent\textit{Data preparation.}
\ourmodel\ requires dual stream and multimodal inputs: TS data $\its$ and FC data $\ifc$.
To get TS data $\its$, we first process the raw fMRI $\ifmri$ by 3 steps:
(1) fMRI pre-processing: $\ifmri$ is processed by professional fMRI pre-processing pipelines (e.g. fMRIPrep~\cite{fmriprep}) with steps like head motion correction.
(2) Resize: The processed fMRI images are resized to match the standard ICBM152 space~\cite{mni}.
(3) Parcellation: we parcelize the resized fMRI data into $N$ regions of interest (ROIs) by an atlas to obtain TS data $\its\in\mathbb{R}^{N\times t}$, where $t$ refers to time steps. 
In this context, an atlas is a neuroscientist-curated annotation that clusters brain voxels based on their functional similarity, and each ROI is a group of neurons that perform similar functions. 
After getting $\its$, we further compute FC data $\ifc$ by the correlation~\cite{cor} between TS data $\its$ and itself along the time dimension $\ifc = \mathrm{Cor}(\its, \its), \ifc\in \mathbb{R}^{N\times N}$.

\noindent\textit{Multimodal learning.}
A key contribution to this work is that we rethink TS and FC as two distinct modalities instead of two features from one modality.
We use two separate encoders: a time-series (TS) encoder $\ets$ and a functional connectivity (FC) encoder $\efc$ to process the multimodal input, which gives us TS and FC features $\rts = \ets(\its)\in\mathbb{R}^M$ and $\rfc = \efc(\ifc) \in\mathbb{R}^M$, respectively. 
Here, $M$ is the size of output features.
To fuse $\rts$ and $\rfc$ and obtain multimodal feature $\rfinal$, a feature fusion scheme  $\mathrm{fusion}(\cdot)$ is introduced: $\rfinal=\mathrm{fusion}(\rts,\rfc)$.

We regard TS and FC as two distinct modalities for two reasons:
(1) Their information is different and complementary.
TS focuses on fine-grained, regional features and provides more temporal information~\cite{ts}, while FC constructs a global brain network emphasizing spatial relationships between regions~\cite{fc}.
(2) For ML models, the global patterns in FC might not be directly learnable from TS data, especially with such small datasets~\cite{mmj1}.
In this situation, FC provides an explicit and complementary global summary of the information in TS data.
To exploit this complementarity, \ourmodel\ treats them as two distinct modalities.

\noindent\textit{FM-powered external knowledge transfer.}
To enrich the limited information in pain-specific data, we use FM to introduce external knowledge.
FM learns rich knowledge by pre-training on large fMRI datasets, with which it can extract robust representations from data and generalize effectively with small, domain-specific pain datasets~\cite{brainlm}.
\ourmodel\ leverages BrainLM~\cite{brainlm} here, which is a foundation model for TS data with rich knowledge obtained from 6,700 hours of fMRI data from 50k+ UK Biobank participants~\cite{ukb2}.

\noindent\textbf{Detailed design of \ourmodel.} We now discuss the designs of encoders $\ets$, $\efc$ and the feature fusion method $\mathrm{fusion}(\cdot)$, and the motivations for design decisions.

\noindent\textit{BrainLM as TS encoder and rationale.}
we use BrainLM's encoder~\cite{brainlm} to implement $\ets$, which is a 4-layer 4-head transformer with an efficient self-attention mechanism to reduce computational complexity while maintaining global context~\cite{nystromformer}. We use the class token of the output of BrainLM to get $\rts$. 
BrainLM is frozen during training to stabilize training and avoid mode collapse when fine-tuned on small, domain-specific pain datasets. Traditional knowledge distillation ~\cite{transfer} is not applied in our study due to
a) limited pain-specific training data, and
b) restricted access to BrainLM's original pre-training dataset: UK Biobank~\cite{ukb2}.

\noindent\textit{ResNet-18 as FC encoder and rationale.}
We simply implement $\efc$ with the backbone of ResNet-18 $E_{ResNet}$~\cite{resnet} alongside a linear projection layer $h$ afterward to align $E_{ResNet}$'s output dimensions with $\rts$. 
Thus, $\efc=h\circ E_{ResNet}$ and we train it from scratch.
The choice of using ResNet for $\efc$ is based on three reasons:
(1) FC matrices capture spatial relationships between brain regions, sharing structural similarities with 2D images. 
(2) While more advanced models like transformers~\cite{bnt} work well in large-scale fMRI analysis, they deliver unsatisfactory performances with limited pain-specific data.
(3) CNN is proven effective for FC data~\cite{brainnetcnn,ai4pain-kevin2}. 
By applying ResNet as $\efc$, we balance simplicity and performance under severe data scarcity.

\noindent\textit{Multimodal feature fusion via concatenation and rationale.}
We apply concatenation ($\mathrm{concat(\cdot)}$), a simple yet effective method to get the fused multimodal feature $\rfinal$ by $\rfinal=\mathrm{concat}(\rts,\rfc), \rfinal\in\mathbb{R}^{2M}$. 
In our preliminary experiments (see Table~\ref{tab:ff}), we found advanced multimodal feature fusion techniques including attention-based~\cite{bi-attention} and mixture-of-experts (MoE)~\cite{moe} delivered unsatisfactory performance, likely due to limited training data.
Element-wise summation's suboptimal performance could be incompatible element-wise semantics in features from the transformer ($\ets$) and ResNet ($\efc$).
Concatenation preserves individual semantics in $\rts$ and $\rfc$, allowing the following layers to dynamically adjust feature prioritization.

\noindent\textbf{Model training.}
\ourmodel\ is trained end-to-end. 
After feature fusion, a single-layer perceptron classifier $p(\cdot)$ is attached for the downstream drug response prediction: $\hat{y}=p(\mathrm{concat}(\ets(\its),\efc(\ifc)))$.
The training is driven by a binary cross-entropy loss between predictions $\mathbf{\hat{y}}$ and labels $\mathbf{y}$: $\mathcal{L}_{BCE}(\mathbf{\hat{y}}, \mathbf{y})$.

\section{Experiments}

\subsection{Experiment Setup}

\begin{table}[t]
\centering
\caption{Simple feature concatenation outperforms advanced feature fusion schemes.}
\label{tab:ff}
\resizebox{0.85\textwidth}{!}{%
\begin{tabular}{c|ccccc}
\hline
\multirow{2}{*}{Method}&  \multirow{2}{*}{\shortstack{Unidirectional\\cross-attention~\cite{bi-attention}}} & \multirow{2}{*}{\shortstack{Bidirectional\\cross-attention~\cite{bi-attention}}}& \multirow{2}{*}{\shortstack{Mixture-of-\\experts~\cite{moe}}}& \multirow{2}{*}{\shortstack{Element-wise\\summation}}& \multirow{2}{*}{Concatenation (ours)}\\
\\
\hline
MCC (\%) & 0 & 44.12 & 0 & 53.93 & \textbf{69.64} \\ \hline
\end{tabular}%
}
\end{table}

\begin{table}[t]
\centering
\caption{Metadata of the in-house dataset and the OpenNeuro~\cite{openneuro} dataset.}
\label{tab:metadata}
\resizebox{\textwidth}{!}{%
\begin{tabular}{c|c|c|c|c|c}
\hline
Dataset & No. & Mean age & Gender & Given drugs & Response labels \\ \hline
In-house & 61 & 58.3 (40-82) &  Male(38), Female(23) & Lidocaine(61) & $y=1$(24), $y=0$(37)\\
OpenNeuro~\cite{openneuro} & 56 & 57.9 (44-73) & Male(26), Female(30) & Duloxetine(19),Placebo(37) & $y=1$(26), $y=0$(30)\\ \hline
\end{tabular}%
}
\end{table}

\noindent\textbf{Datasets.}
The metadata of the in-house and OpenNeuro~\cite{openneuro} datasets are shown in Table~\ref{tab:metadata}.
The OpenNeuro dataset is the patient cohort of the DS000208 dataset from the OpenNeuro platform~\cite{openneuro}.
The in-house data are collected from the local hospital.
fMRI scans in both datasets are pre-treatment scans.

\noindent\textbf{Baselines.}
We compare a) \ourmodel\ as an end-to-end predictive model against four TS baselines (LSTM~\cite{lstm}, ResNet~\cite{resnet}, BrainLM~\cite{brainlm} w/wo pre-training) and two FC baselines (ResNet~\cite{resnet}, BNT~\cite{bnt}), and b) \ourmodel-extracted feature $\rfinal$ for linear machine learning classifiers against raw TS/FC data ($\its$, $\ifc$), PCA-extracted features (PCA($\its$), PCA($\ifc$)), and BrainLM-derived features (BrainLM($\its$))~\cite{brainlm}. 
We also provide a random guessing experiment simulating conventional random participant recruitment in clinical trials.
Random guessing is repeated 1,000 times.
Other reported results are under 5-fold cross-validation. 

\noindent\textbf{Implementation of \ourmodel.}
We use fMRIPrep~\cite{fmriprep} in fMRI pre-processing. We use A424~\cite{a424} atlas in parcellation following BrainLM~\cite{brainlm}.
The number of ROIs $N$ is 424.
$\its$ is truncated or padded to a fixed length of 200 and patchified with a patch size of 20.
The feature dimension $M$ is 256.

\noindent\textbf{Metrics.}
We report four metrics: F1, balanced accuracy (BACC), AUROC, and Matthew correlation coefficient (MCC).
Because MCC fairly measures TP, TN, FP, and FN, we use MCC as the primary metric to select the best models.
The bests are \textbf{bolded} and the second bests are \underline{underlined}.

\noindent\textbf{Experiment design of drug response prediction.}
We perform drug-agnostic and drug-specific response prediction in experiments.
\emph{Drug-agnostic response prediction} is to predict whether a patient is a responder regardless of the types of drugs and can be mathematically described as $\hat{y}=f_\mathbf{\bm{\theta}}(\mathbf{X}_{raw})$.
It is the first step in clinical trials, reducing trial costs and time by focusing on likely responders~\cite{treatment2}.
\emph{Drug-specific response prediction} is to predict the responses for specific drugs and can be described as $\hat{y}=f_\mathbf{\bm{\theta}}(\mathbf{X}_{raw}|d)$, where $d$ is the type of given drugs.
It can help clinicians select participants that are more targeted and fine-grained.

\subsection{Experiment Results}\label{sec:results}

\begin{table}[t!]
\centering
\caption{\ourmodel\ vs. unimodal baselines on drug-agnostic response prediction. The results of the ablation study are also shown by comparing Exp. 5, 7, 8, 9.}
\label{tab:agnostic}
\resizebox{\textwidth}{!}{
\begin{threeparttable}
\begin{tabular}{ccccccccc}
\hline 
\multirow{2}{*}{Dataset} & \multirow{2}{*}{\shortstack{Exp.\\ID}} & \multicolumn{1}{c}{\multirow{2}{*}{Model}} & \multirow{2}{*}{\shortstack{Input\\modality}} & \multirow{2}{*}{\shortstack{Pre-\\training}} & \multicolumn{4}{c}{Metrics (\%)} \\ \cline{6-9} 
 & \multicolumn{1}{c}{} & & &  & F1$\uparrow$ & BACC$\uparrow$ & AUROC$\uparrow$ & MCC$\uparrow$ \\ 
\hline 
\multirow{9}{*}{\rotatebox{90}{OpenNeuro~\cite{openneuro}}} & 1 & Random & N/A & \xmark & 49.57$\pm$11.39 & 50.64$\pm$11.49 & 50.64$\pm$11.49 & 1.29$\pm$22.96 \\
 & 2 & LSTM~\cite{lstm} & TS & \xmark & 63.95$\pm$11.94 & 67.33$\pm$8.38 & 51.67$\pm$12.80 & 43.68$\pm$16.89 \\
 & 3 & BNT~\cite{bnt} & FC& \xmark & 34.90$\pm$0.88 & 50.00$\pm$0.00 & 51.00$\pm$18.88 & 0.00$\pm$0.00 \\
 & 4 & ResNet~\cite{resnet} & TS& \xmark & 63.36$\pm$9.21 & 67.00$\pm$7.58 & 50.00$\pm$14.91 & 42.79$\pm$12.90 \\
 & 5 & ResNet~\cite{resnet} & FC& \xmark & 66.79$\pm$12.64 & 69.33$\pm$10.97 & 55.11$\pm$29.58 & 44.23$\pm$20.91 \\
 & 6 & BrainLM~\cite{brainlm} & TS& \xmark & 34.90$\pm$0.88 & 50.00$\pm$0.00 & 44.56$\pm$16.50 & 0.00$\pm$0.00 \\
 & 7 & BrainLM~\cite{brainlm} & TS& \cmark & \underline{79.38$\pm$8.77} & \underline{80.00$\pm$8.16} & \underline{71.44$\pm$15.28} & \underline{63.36$\pm$14.47} \\
 & 8 & \ourmodel & TS+FC& \xmark & 70.18$\pm$14.04 & 73.67$\pm$11.87 & 60.67$\pm$25.43 & 54.93$\pm$18.50 \\
 & 9 & \ourmodel & TS+FC& \cmark & \textbf{80.51$\pm$16.11} & \textbf{82.33$\pm$14.12} & \textbf{77.56$\pm$26.98} & \textbf{69.64$\pm$23.32} \\
\hline 
\multirow{9}{*}{\rotatebox{90}{In-house}} & 1 &  Random & N/A & \xmark & 47.96$\pm$11.61 & 49.12$\pm$11.66 & 49.12$\pm$11.66 & -1.77$\pm$23.35 \\
 & 2 & LSTM~\cite{lstm} & TS& \xmark & 37.72$\pm$1.38 & 50.00$\pm$0.00 & 64.84$\pm$37.41 & 0.00$\pm$0.00 \\
 & 3 & BNT~\cite{bnt} & FC& \xmark & 34.90$\pm$0.88 & 50.00$\pm$0.00 & 51.00$\pm$18.88 & 0.00$\pm$0.00 \\
 & 4 & ResNet~\cite{resnet} & TS& \xmark & 66.98$\pm$12.74 & 68.67$\pm$11.63 & 59.67$\pm$15.20 & 41.86$\pm$22.42 \\
 & 5 & ResNet~\cite{resnet} & FC& \xmark & \underline{75.68$\pm$9.33} & 75.64$\pm$9.27 & 68.96$\pm$11.74 & \underline{60.92$\pm$12.58} \\
 & 6 & BrainLM~\cite{brainlm} & TS& \xmark & 34.90$\pm$0.88 & 50.00$\pm$0.00 & 44.56$\pm$16.50 & 0.00$\pm$0.00 \\
 & 7 & BrainLM~\cite{brainlm} & TS& \cmark & 74.83$\pm$16.81 & \underline{75.71$\pm$15.32} & \underline{73.88$\pm$24.06} & 59.01$\pm$26.53 \\
 & 8 & \ourmodel & TS+FC& \xmark & 66.92$\pm$21.59 & 72.14$\pm$16.95 & 71.43$\pm$21.29 & 48.32$\pm$32.64 \\
 & 9 & \ourmodel & TS+FC& \cmark & \textbf{80.32$\pm$13.89} & \textbf{81.07$\pm$13.87} & \textbf{75.20$\pm$26.14} & \textbf{63.72$\pm$26.77}\\
\hline 
\end{tabular}
\end{threeparttable}
}
\end{table}

\noindent\textbf{Superiority of \ourmodel\ in end-to-end prediction.}
We evaluate \ourmodel\ on \emph{drug-agnostic response prediction} against:
a) four baselines for TS data (LSTM~\cite{lstm}, ResNet~\cite{resnet}, BrainLM~\cite{brainlm} w/wo pre-training) and 
b) two baselines for FC data (ResNet~\cite{resnet}, BNT~\cite{bnt}) on OpenNeuro and in-house datasets (see Table~\ref{tab:agnostic}). 
BrainLM without pre-training (Exp. 6) and BNT (Exp. 3) struggle with limited training data, highlighting challenges in training advanced ML with small fMRI datasets. 
Other unimodal baselines (Exp. 2, 4, 5, 7) show inconsistent results across datasets. 
\ourmodel\ consistently outperforms all baselines, demonstrating superior prediction performance and cross-dataset adaptability.

\begin{table}[t!]
\centering
\caption{We compare the representation ability (within-domain) and generalizability (out-of-domain) of \ourmodel\ and BrainLM on drug-specific response prediction using the OpenNeuro dataset. Given BrainLM's expected generalizability as a foundation model and its rank as the second-best model on the OpenNeuro dataset in Table~\ref{tab:agnostic}, we focus our comparison on \ourmodel\ and BrainLM.}
\label{tab:specific}
\resizebox{\textwidth}{!}{
\begin{threeparttable}
\begin{tabular}{cccccccc}
\hline
  \multirow{2}{*}{Setup}&  \multirow{2}{*}{\shortstack{Training\\drug}} & \multirow{2}{*}{\shortstack{Testing\\drug}}&\multicolumn{1}{c}{\multirow{2}{*}{Model}} & \multicolumn{4}{c}{Metrics (\%)} \\ \cline{5-8} 
 & \multicolumn{1}{c}{} & &  & F1$\uparrow$ & BACC$\uparrow$ & AUROC$\uparrow$ & MCC$\uparrow$ \\ 
\hline 
\multirow{4}{*}{\shortstack{Within-\\domain}} & \multirow{2}{*}{Duloxetine}& Duloxetine & BrainLM~\cite{brainlm}  & 81.33$\pm$29.21 & 85.00$\pm$22.36 & \textbf{90.00$\pm$22.36} & 71.55$\pm$43.98 \\
 & & Duloxetine & \ourmodel & \textbf{84.00$\pm$14.60} & \textbf{86.67$\pm$12.63} & 85.00$\pm$23.36 & \textbf{74.64$\pm$23.14}  \\
 \cline{2-8}
 & \multirow{2}{*}{Placebo}& Placebo & BrainLM~\cite{brainlm}  & 77.84$\pm$12.87 & 79.17$\pm$12.84 & 78.33$\pm$21.17 & 60.37$\pm$25.68 \\
 & & Placebo & \ourmodel & \textbf{88.95$\pm$6.29} & \textbf{89.17$\pm$6.32} & \textbf{89.17$\pm$9.93} & \textbf{80.59$\pm$11.00}  \\
\hline
\multirow{4}{*}{\shortstack{Out-of-\\domain}} & \multirow{2}{*}{Duloxetine}& Placebo & BrainLM~\cite{brainlm}  & 55.40$\pm$16.43 & 58.19$\pm$13.43 & 71.36$\pm$9.21 & 16.67$\pm$27.20 \\
 & & Placebo & \ourmodel & \textbf{92.30$\pm$4.97} & \textbf{92.27$\pm$5.41} & \textbf{91.59$\pm$7.73} & \textbf{86.02$\pm$8.91}  \\
 \cline{2-8}
 & \multirow{2}{*}{Placebo}& Duloxetine & BrainLM~\cite{brainlm}  & 52.58$\pm$6.00 & 56.61$\pm$4.91 & 66.02$\pm$2.32 & 17.26$\pm$13.51 \\
 & & Duloxetine & \ourmodel & \textbf{83.24$\pm$14.93} & \textbf{84.59$\pm$12.20} & \textbf{89.77$\pm$9.93} & \textbf{71.85$\pm$19.25}  \\
\hline 
\end{tabular}
\end{threeparttable}
}
\end{table}

We evaluate \ourmodel\ on \emph{drug-specific response prediction} against BrainLM~\cite{brainlm} (see Table~\ref{tab:specific}).
We divide the OpenNeuro dataset into two subsets according to the given drugs.
We fine-tune BrainLM~\cite{brainlm} and train \ourmodel\ on one subset and test them on both subsets.
Results of \emph{within-domain} tests show that \ourmodel\ outperforms BrainLM when training and testing on the same drug, illustrating \ourmodel's superior representation ability.
\ourmodel's strong generalizability is demonstrated by \emph{out-of-domain} evaluation: 
when training with one drug and transferring to another, \ourmodel\ outperforms BrainLM by a significant margin.

\noindent\textbf{Effectiveness of multimodal feature $\rfinal$ for linear classifiers.}
We evaluate \ourmodel's representation ability as a feature extractor for linear machine learning classifiers on \emph{drug-agnostic response prediction} with the OpenNeuro dataset.
We compare multimodal feature $\rfinal$ from \ourmodel\ against unimodal features from raw data, PCA, and BrainLM~\cite{brainlm} with four commonly used linear classifiers~\cite{ai4pain-review1}: SVM, Ridge, k-NN, and XGBoost.
The results in Fig.~\ref{fig:intra-dataset}(a) show that statistical machine learning approaches struggle to capture the complex patterns in fMRI with their limited capacity.
In contrast, \ourmodel\ outperforms all other feature extractors by a significant margin, indicating the effectiveness of multimodal feature $\rfinal$ and the strong representation ability of \ourmodel.

\noindent\textbf{Ablation study.}
We now demonstrate the effectiveness of multimodal learning and FM-powered external knowledge transfer (see Exp. 5, 7, 8, 9 in Table~\ref{tab:agnostic}).

\noindent\textit{Multimodal learning} improves MCC by at least \textbf{2.80\%}.
The effectiveness of multimodal learning is demonstrated by comparing Exp. 5, 7, and 9 in Table~\ref{tab:agnostic}, where ResNet with FC and BrainLM with TS serve as two separate streams of \ourmodel. 
The results indicate that the performance of ResNet (Exp. 5) and BrainLM (Exp. 7) varies across datasets, whereas \ourmodel\ (Exp. 9) consistently outperforms them, achieving strong and stable performance across datasets. 
This highlights the superiority and cross-dataset adaptability of multimodal learning.

\noindent\textit{FM-powered external knowledge transfer} improves MCC by at least \textbf{14.71\%}.
We evaluate the effectiveness of external knowledge in \ourmodel\ by comparing \ourmodel\ w/wo pre-training (Exp. 8, 9 in Table~\ref{tab:agnostic}).
The performance gap between the two models (69.64\%/63.72\% vs. 54.93\%/48.32\% on MCC) confirms the importance of external knowledge and pre-training to address data scarcity.

\begin{figure}[t!]
    \centering
    \includegraphics[width=\linewidth]{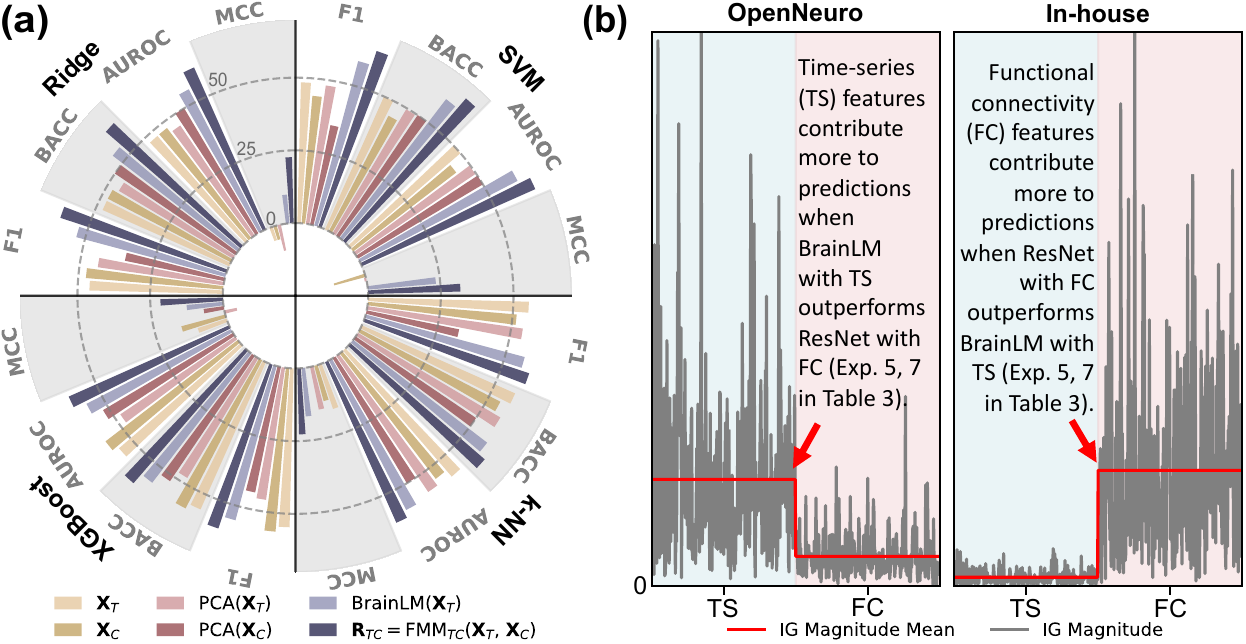}
    \caption{
    \textbf{(a)} Multimodal feature $\rfinal$ from \ourmodel\ outperforms unimodal features from other feature extractors with linear classifiers on drug-agnostic response prediction.
    \textbf{(b)} Feature importance via integrated gradients (IG) illustrates how \ourmodel's cross-dataset dynamic behaviors enhance adaptability: \ourmodel\ flexibly prioritizes the most effective modality in predictions.
    The light blue background is IG values for time-series feature $\rts$ and the light red region is for functional connectivity feature $\rfc$.
    }
    \label{fig:intra-dataset}
\end{figure}

\noindent\textbf{Interpretation to \ourmodel's cross-dataset adaptability.}\label{sec:interp}
The high adaptability of \ourmodel\ is due to its flexible dynamic behaviors.
We use integrated gradients (IG)~\cite{ig} to explain how \ourmodel's flexible reliance on multimodal features enhances its adaptability across datasets. 
A larger IG magnitude signifies a higher impact of input features on the model's predictions. 
We present IG magnitudes for the final predictor of \ourmodel, which takes the fused multimodal features $\rfinal$ as inputs and outputs predictions $\mathbf{\hat{y}}$, shown in Fig.~\ref{fig:intra-dataset}(b). 

On the OpenNeuro dataset, TS features contribute more to the predictions than FC features, while FC features have a stronger impact on the in-house dataset.
This observation aligns with the results of Exp. 5 and 7 in Table~\ref{tab:agnostic}. 
When BrainLM with TS outperforms ResNet with FC on the OpenNeuro dataset, \ourmodel\ relies more on the TS features and capitalizes on the strengths of TS data.
In the in-house dataset, ResNet with FC performs better than BrainLM with TS, causing \ourmodel\ to emphasize FC features in prediction.

These findings highlight \ourmodel's cross-dataset dynamic behaviors and explain \ourmodel's high adaptability.
By combining TS and FC modalities, \ourmodel\ can dynamically adjust its reliance on various modalities based on the dataset and task. 
This flexibility enables \ourmodel\ to adapt effectively across diverse conditions and data distributions, improving performance in various scenarios.

\section{Conclusion}

To address the severe data scarcity in fMRI-based neuropathic pain drug response prediction, we propose \ourmodel, a foundation-model-boosted multimodal learning framework with multimodal learning and foundation-model-powered external knowledge transfer.
\ourmodel\ shows strong representation ability, generalization, and cross-dataset adaptability, enabling \ourmodel\ with abilities to predict treatment effects and enhance stratification in the participant recruitment process, and with the potential to significantly reduce the time and money costs associated with random methods in traditional clinical trials.

\begin{credits}
\noindent\textbf{\ackname} 
This study is funded by the Engineering and Physical Sciences Research Council (EP/Y017544/1) for the "A Novel Artificial Intelligence Powered Neuroimaging Biomarker for Chronic Pain" project.

\end{credits}

\bibliographystyle{splncs04}
\bibliography{reference}
\end{document}